# THE IDENTIFICATION AND CATEGORIZATION OF ANEMIA THROUGH ARTIFICIAL NEURAL NETWORKS: A COMPARATIVE ANALYSIS OF THREE MODELS

Mohammed A. A. Elmaleeh

Computer Engineering Department, Faculty of Computer and Information Technology, Tabuk, Saudi Arabia

## ABSTRACT

*This paper presents different neural network-based classifier algorithms for diagnosing and classifying Anemia. The study compares these classifiers with established models such as Feed Forward Neural Network (FFNN), Elman network, and Non-linear Auto-Regressive Exogenous model (NARX). Experimental evaluations were conducted using data from clinical laboratory test results for 230 patients. The proposed neural network features nine inputs (age, gender, RBC, HGB, HCT, MCV, MCH, MCHC, WBCs) and one output. The simulation outcomes for diverse patients demonstrate that the suggested artificial neural network rapidly and accurately detects the presence of the disease. Consequently, the network could be seamlessly integrated into clinical laboratories for automatic generation of Anemia patients' reports Additionally, the suggested method is affordable and can be deployed on hardware at low costs.*

## 1. INTRODUCTION

Blood is a crucial and intricate fluid that flows through the body's blood vessels, supporting essential bodily functions in organisms, which includes humans. Composed of both liquid and cellular components, blood plays a crucial role in transporting oxygen, nutrients, hormones, and waste products throughout the body. It composes of various components, including red blood cells (RBC), hemoglobin (HGB), hematocrit (HCT), mean corpuscular volume (MCV), mean corpuscular hemoglobin (MCH), mean corpuscular hemoglobin concentration (MCHC), and white blood cells (WBCs). These components are essential for various physiological functions, and analyzing them through clinical laboratory tests can provide valuable information for diagnosing and classifying conditions such as Anemia [1, 2].

### 1.1. Anemia

Anemia is a medical condition characterized by a deficiency of red blood cells (RBCs) or a decreased amount of hemoglobin in the blood. Hemoglobin is a protein in red blood cells that binds with oxygen and helps transport it throughout the body. Anemia can result in reduced oxygen-carrying capacity, leading to various symptoms and potential health complications. The common causes of Anemia include: Vitamin Deficiencies, Iron Deficiency, Chronic Infections or Inflammation, Genetic Disorders and Chronic Diseases. Anemia typically manifests with prevalent symptoms such as fatigue, weakness, pale skin, shortness of breath, dizziness, and headaches. The appropriate treatment for Anemia varies depending on the root cause and may encompass dietary adjustments, supplementation with iron or vitamins, prescribed medications, or other necessary interventions [2].

### 1.1.1. Blood Testing

Blood testing is a medical procedure involving the collection and analysis of a blood sample to assess various aspects of a person's health. This diagnostic method provides valuable information about the composition of blood, including the levels of different cells, proteins, hormones, and other substances.

### 1.1.2. Blood Count

A complete blood count is a set of examinations employed to assess the cellular constituents and concentration within the bloodstream. These tests include assessing red blood cell count (RBC), red cell distribution width (RDW), white blood cell count (WBC), mean corpuscular hemoglobin (MCH), mean cell hemoglobin concentration (MCHC), hemoglobin levels (HGB), hematocrit (HCT), and mean cell volume (MCV) [3].

Doctors today utilize hologram blood test results to diagnose Anemia based on the data obtained. The parameters examined in this study for identifying the disease include [4]:

- If MCV, MCH, and MCHC show a decrease, it suggests microcytic, hypochromic Anemia, often linked to iron deficiency
- Elevated MCV, MCHC, and MCH values point to macrocytic Anemia, typically caused by deficiencies in vitamin B12 and folic acid
- When MCV, MCH, and MCHC levels are within normal ranges, it indicates normocytic,
- normochromic Anemia, commonly associated with acute blood loss.

In Sudan, most hospitals and healthcare facilities still use manual methods to analyse patient data concerning Anemia. This method depends on the expertise of clinicians, resulting in time-consuming processes and potential stress for patients. To enhance this system, a designed neural network incorporates various classifier algorithms to diagnose and classify different types of Anemia. This system compares classifiers such as feed-forward neural network (FFNN), Elman network, and Non-linear Auto-Regressive exogenous model (NARX) to achieve heightened accuracy in the diagnosis and classification of Anemia.

Upon reviewing literature pertaining to the classification of Anemia, it is evident that a diverse range of methods has been employed. For instance, a fuzzy expert system has been enhanced to diagnose Sickle Cell Anemia. Fuzzy Systems have also been applied to diagnose different Anemia types using symptoms like Irritability, Tachycardia, Memory Weakness, Bleeding, and Chronic Fatigue [5, 6].

The study on diagnosing iron-deficiency anemia in women uses a multi-model neural network approach with input data including RBC, HGB, HCT, MCV, MCH, and MCHC. Different neural network methods including Feed Forward Neural Network, Long Short-Term Memory, Elman networks, Non-linear Auto-Regressive exogenous model, and Gated Recurrent Unit are used and compared for effectiveness. Additionally, two neural network models using ZPP, HB, RBC, and MCV inputs are employed for diagnosing anemia, and their performances are compared. [7, 8].

Electrical and Electronics Engineering: An International Journal (ELELIJ) Vol.13, No.1/2, May 2024

## 1.2. Type of Neural Network Models

There are various types of neural network models, each designed for specific tasks and structured in different ways. Some common types of neural network models include:

### 1.2.1. Feed Forward Neural Network (FFNN)

A Feedforward Neural Network (FFNN) is a type of artificial neural network where information flows in one direction, from the input layer to the output layer, without any feedback loops. It's the most basic neural network, with neuron connections moving only forward. FFNN consists of Input Layer, Hidden Layers, Output Layer, Weights, Biases, and Activation Functions. Sometimes referred to as Multi-layered Networks of Neurons (MLN), FFNNs use supervised learning algorithms like backpropagation to adjust weights and biases based on predicted and actual outputs. They find extensive use in pattern recognition, image, and speech recognition, classification, regression, and other machine learning tasks. Figure1 depicts a model of multilayer feed forward back propagation neural network [9].

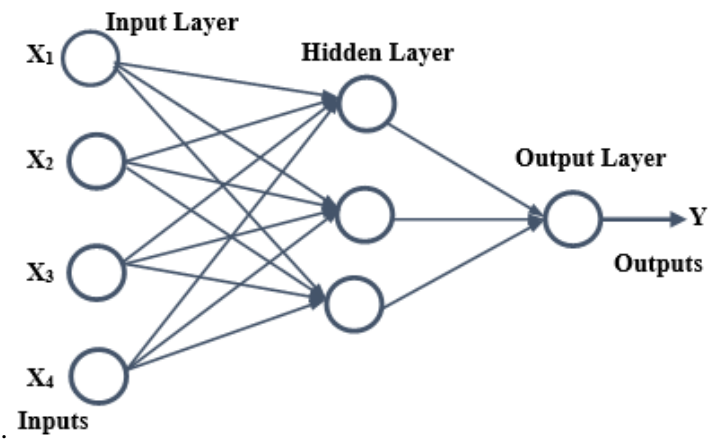

Figure 1: A model of multilayer feed forward back propagation neural network [9].

The network's activation function will be applied the sigmoid activation function [9].

### 1.2.2. Long Short-Term Memory (LSTM)

Long Short-Term Memory (LSTM) networks are a type of recurrent neural network (RNN) architecture designed to address the challenges of capturing and learning long-term dependencies in sequential data. Developed to overcome the limitations of traditional RNNs, LSTMs have proven effective in various applications, particularly those involving time series data, natural language processing, and speech recognition [10]. LSTM, developed by Hochreiter and Schmidhuber, addresses issues in traditional RNNs and machine learning algorithms. It can be implemented in Python using the Keras library. Figure2 shows the Structure of an LSTM Network.

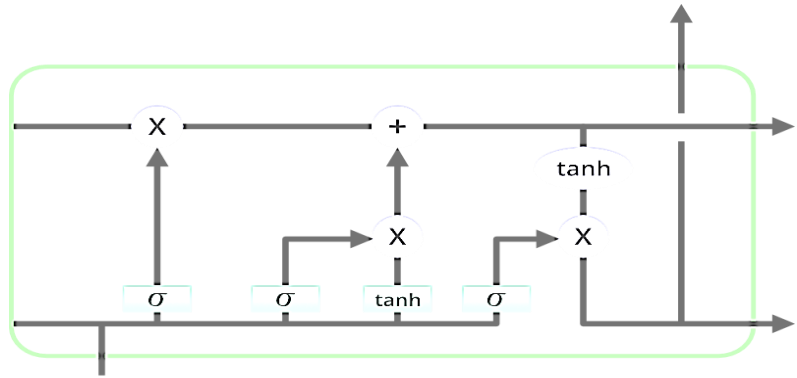

Figure2: Structure of an LSTM Network

### 1.2.3. Gated Recurrent Unit (GRU)

The Gated Recurrent Unit (GRU) is a type of recurrent neural network (RNN) designed to handle dependencies in sequential data, similar to LSTM networks. Both GRUs and LSTMs address the vanishing gradient problem in traditional RNNs, allowing them to learn and retain information effectively over long sequences. While they share similarities, the choice between GRUs and LSTMs depends on the task's requirements. GRUs may be preferred for computational efficiency, as they offer a slightly simpler model structure without sacrificing performance.

### 1.2.4. Elman Networks Model

Elman Neural Networks, named after Jeffrey Elman, are a type of recurrent neural network (RNN) designed to capture temporal dependencies in sequential data. They feature hidden layers with recurrent connections, enabling short-term memory. Elman networks excel in tasks like time series prediction and language modelling due to their ability to understand sequential patterns. Although simpler than LSTM networks, Elman networks remain effective for applications requiring understanding of sequential relationships. They are applied in fields such as speech recognition and system identification, demonstrating versatility in learning temporal dependencies across various data types. Figure3 illustrates the common structure of an Elman Network [10].

Elman's network and various other neural network models were designed with the purpose of configuring input and output units to correspond to individual letters. This configuration facilitates the training of the network to anticipate and predict the subsequent letter in a given sequence of letters.

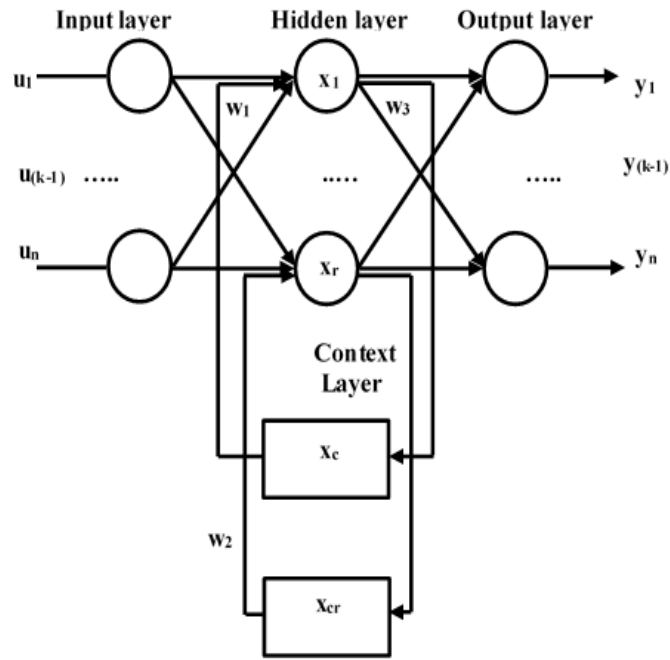

Figure3: A common structure of an Elman network [10]

**1.2.5. NARX Model**

The Non-linear Auto-Regressive with exogenous inputs (NARX) model is a neural network architecture utilized for modelling and forecasting time-series data. It incorporates both autoregressive and exogenous inputs to predict future values based on past observations and external factors. NARX is extensively utilized in the fields of control systems and signal processing, demonstrating proficiency in forecasting future values by considering historical patterns and external factors. It serves as a versatile instrument for analyzing time-series data. Figure4 depicts a model of the NARX [10].

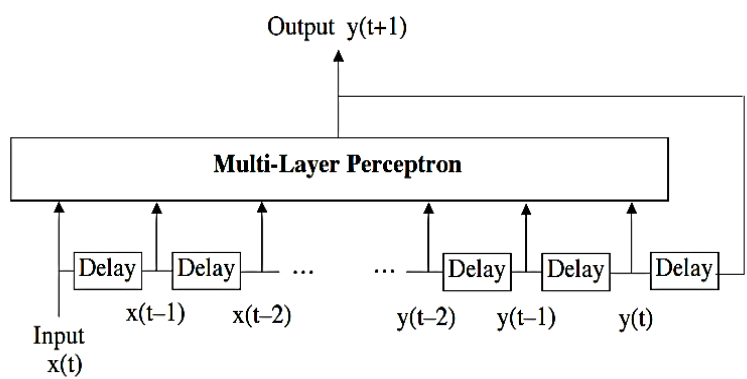

Figure4: A common structure of NARX network [11].

This model demonstrates notable efficacy in Time Series Prediction, specifically when the goal is to predict the value of x at time t+1 (denoted as y(t+1)). In algebraic terms, the model can be expressed as predicting the next instance (t+1) of x given the current instance (t).

$$y(t) = F(X_{t-1}, X_{t-2}, X_{t-3}, \ldots, Y_t, Y_{t-1}, Y_{t-2}, \ldots) + \sum \epsilon \qquad (1)$$

In this context, 'y' represents the variable under consideration, while 'u' signifies the externally influenced variable. In this context, data represented by 'u' acts as a forecaster for 'y,' and this forecast is additionally impacted by previous values of 'y.' The symbol 'ε' is introduced as the residual term, sometimes denoted as the disturbance in the system.

## 2. METHODS IMPLEMENTED

The design of the neural network model proposed includes nine neurons in the input layer, a hidden layer containing one hundred neurons, and one neuron in the output layer. The activation function utilized within the network is the sigmoid function. Learning is achieved through gradient descent with momentum for weight and bias adjustment [11].

The methodology employed for the diagnosis and classification of Anemia disease is structured into three distinct modules:

1. Pre-processing of the input data
2. The application, training, and evaluation of the selected Artificial Neural Network (ANN) model.
3. Post-processing of the output data.

The detailed partitioning of the second module into distinct diagnosis and classification sub-components, as depicted in Figure 4, serving as an illustrative depiction of the block diagram for the proposed approach. The pre-processed data undergoes initial scrutiny to determine the existence of the disease in the test subjects. In cases of positive diagnoses, the outcomes are subsequently transmitted to the classification sub-module, which identifies the precise category and type of Anemia. This methodological framework ensures a systematic and rigorous analysis during both the diagnostic and classification procedures.

### 2.1. Data Pre-Processing

For this study, data used to train the neural network were sourced from Omdurman Teaching Hospital in Sudan. The dataset consisted of 230 samples, comprising both Anemia and non-Anemia cases. The Anemic cases are further categorized into three subtypes: microcytic, normocytic, and macrocytic. Clinical test results undergo thorough examination and organization. Following a comprehensive consultation with domain experts, seven crucial tests are identified for the diagnosis and classification of Anemia disease. These test outcomes, pertaining to different patients, are structured into vectors, each comprising seven rows. Targets for each patient are designated as either healthy or unhealthy. This procedure is iterated for both the training and test sets. The network's variables are subsequently defined, forming the basis for input and target data fed into the proposed model. The discussion covers the seven selected tests, as well as the training and test sets employed for the diagnostic and classification processes.

### 2.1.1. Training Set

The dataset used for training consists of 147 samples, encompassing individuals with and without Anemia. Among these, 105 individuals are identified as Anemia-positive, while 42 individuals are categorized as Anemia-negative. Within the Anemia-positive group, there are three subcategories. Group two comprises 26 instances of microcytic Anemia, group three includes 40 instances of normocytic Anemia, and group four consists of 39 instances of macrocytic Anemia

### 2.1.2. Test Dataset

The test dataset comprises 83 samples, and similar to the training set, it is applied to all three networks. For the test set, only the inputs are presented to the network, as the network model has already undergone training and can classify the input into one of the specified classes (microcytic, normocytic, macrocytic, and non-Anemia).

The neural network toolbox within the MATLAB environment is utilized for the creation, training, and testing of the network. Initially, data normalization is performed, scaling the training and test data between -1 and 1. Subsequently, variables are assigned to training, testing, and output data to facilitate their utilization. The interface of the Neural Network Toolbox is depicted in Figure 5.

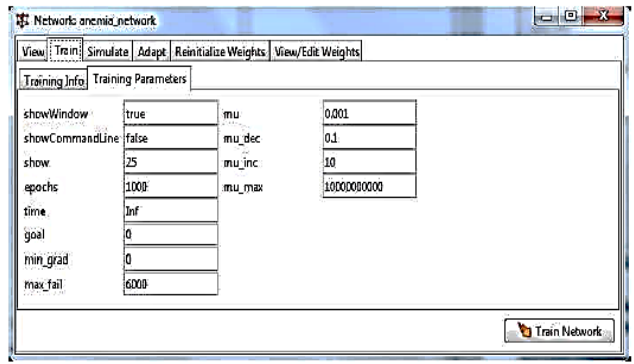

Figure 5: Neural Network Toolbox interface

The input data for both training and testing phases are conveyed, with the output from the training data serving as the target data. The network is constructed using the "Create Network or Data" interface, as illustrated in Figure6.

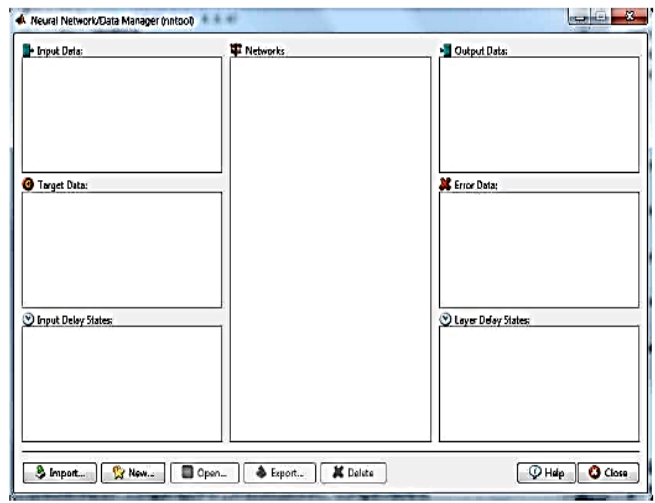

Figure6: Create Network or Data interface

## 2.2. ANN Processing

ANN processing refers to the operations or computations carried out by an Artificial Neural Network (ANN). This processing involves the manipulation of input data through the network's layers of interconnected neurons to produce an output. The operations performed within the ANN include the aggregation of input signals, the application of activation functions, and the propagation of signals through the network's connections. ANN processing enables the network to learn from input data, make predictions, classify objects, or perform other tasks based on the specific architecture and objectives of the neural network. The ANN processing comprises two distinct sub-modules:

### 2.2.1. Diagnosis

When the pre-processed data are input into the ANN, it initially determines whether the individual is anemic or not. The system utilizes the following inputs: HGB, HCT, MCV, MCH, MCHC, age, and gender. The output is a binary value, where 0 denotes a healthy person, and 1 indicates a person with positive Anemia. Subsequently, the data is forwarded to the next sub-module.

### 2.2.2. Classification

This sub-module categorizes the Anemia stage as microcytic, normocytic, or macrocytic. The outcomes of these modules serve as indicators of the network's successful training. Following effective training, the network moves on to the testing stage, where its performance is evaluated using the results obtained from the test dataset. Conversely, if the network fails to diagnose and classify accurately, it undergoes retraining with variations in the number of layers, neurons, and functions.

## 2.3. Post-Analysis Of the Obtained Data

To present the results of individual test samples, it is essential to conduct post-processing on the output data. The outputs from the output layer of the ANNs are transformed into an interpretable format. These results indicate whether the individual is positive for Anemia and specify the classification type of Anemia. Figure7 illustrates the schematic representation of the of the three phased proposed in this study.

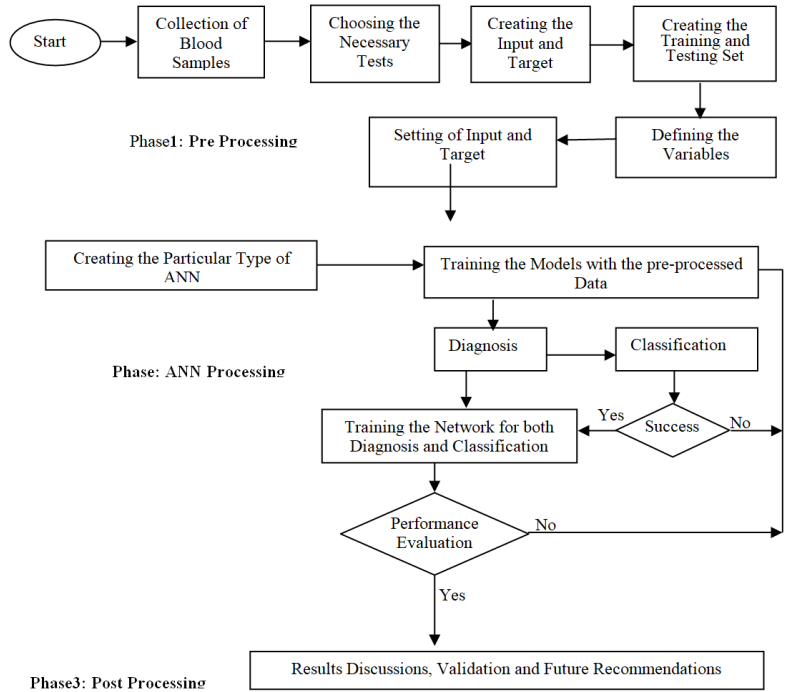

Figure7. The schematic representation of the proposed Methods

## 3. RESULTS AND DISCUSSIONS

The training and testing of neural models for various Anemia detection and classification scenarios were performed using the MATLAB programming environment. The data for each disease were partitioned into 40% for training, 40% for testing, and 20% for validation. The testing dataset was evenly split between the test and validation segments. A single hidden layer was employed in the network structure, with a consistent choice of 50 neurons in the hidden layer across all datasets to ensure result independence across different Anemia types.

In the domain of neural networks, data pertaining to individuals exhibiting normal health, those diagnosed with microcytic Anemia, individuals with normocytic Anemia, and those presenting macrocytic Anemia were systematically and randomly allocated across the learning, validation, and test groups. The findings in Table 1 elucidate that the appropriately configured Feedforward Neural Network (FFNN) demonstrated a precision rate of 87.34%.

Table 1: Comparative Analysis of Accuracy Across Diverse Algorithmic Approaches

| Algorithm Type | Training set | | | |
| --- | --- | --- | --- | --- |
| | Total No of Sampling | Result | | Accuracy |
| | | Positive | Negative | |
| FFNN mode | 230 | 201 | 29 | 87.39% |
| NARX Model | 230 | 209 | 21 | 90.87% |
| Elman Model | 230 | 215 | 15 | 93.48% |

Table2 illustrates the precision, recall (sensitivity), and F1 score values for the three models, with each model's corresponding true positive value. This information is provided based on a total sample size of 230 instances. Overall, all models demonstrate high precision and F1 scores, indicating strong performance in correctly identifying positive instances. Elman Model shows the highest recall, precision, and F1 score among the three models, suggesting superior performance in correctly identifying positive instances while maintaining precision. NARX Model follows closely with slightly lower recall, precision, and F1 score values. FFNN Model exhibits the lowest recall among the three models, but still maintains high precision and F1 score.

Table 1: The precision, recall (sensitivity), and F1 score values for the three models

| Algorithm Type | Total Number of Samples | Recall | Precision | F1 Score |
|---|---|---|---|---|
| FFNN Model | 230 | 0.8739 | 1.0 | 0.9323 |
| NARX Model | 230 | 0.9087 | 1.0 | 0.9522 |
| Elman Model | 230 | 0.9348 | 1.0 | 0.9663 |

Conversely, the Nonlinear Autoregressive with Exogenous Inputs (NARX) model yielded a precise outcome of 90.86% for the test set, while the Elman model exhibited a commendable accuracy of 93.99% for the test set under similar sample conditions. Figures 8-12 delineate the instructional efficacy of the FFNN network, the instructive efficacy of the NARX network, the instructive efficacy of the Elman network, regression outcomes, and training state results, respectively.

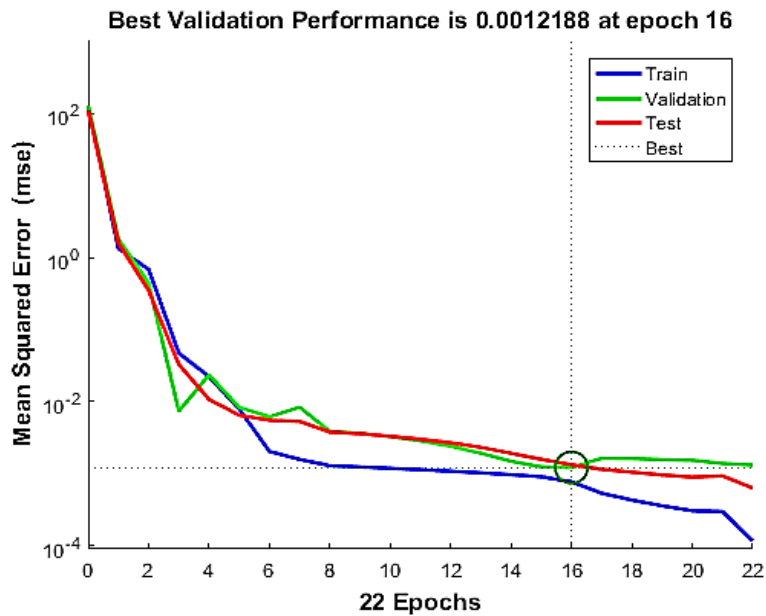

Figure8: Learning Performance of FFNN Network

Electrical and Electronics Engineering: An International Journal (ELELIJ) Vol.13, No.1/2, May 2024

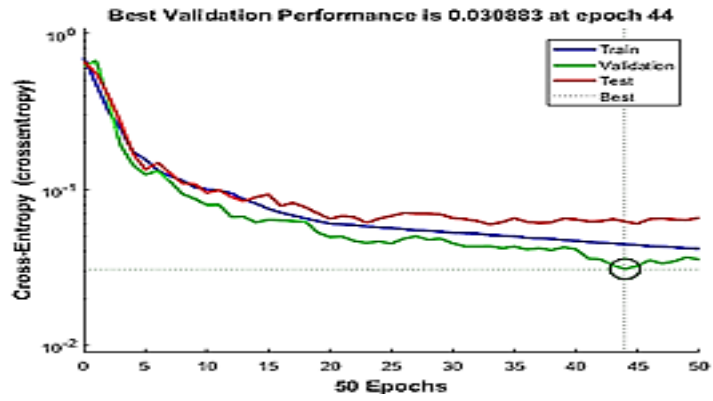

Figure 9: Learning Performance of NAXR Network

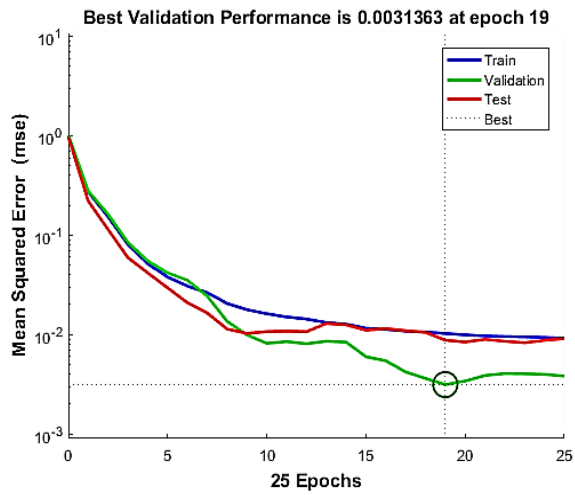

Figure 10: Learning Performance of Elman Network

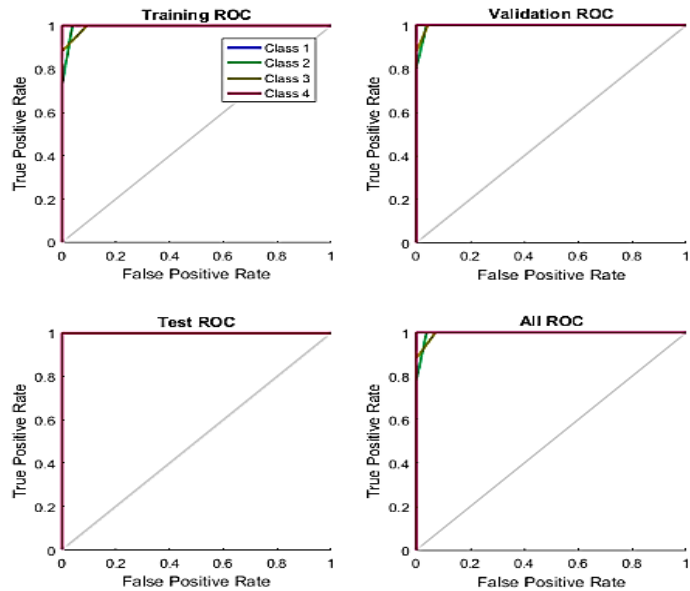

Figure11: Regressions Results

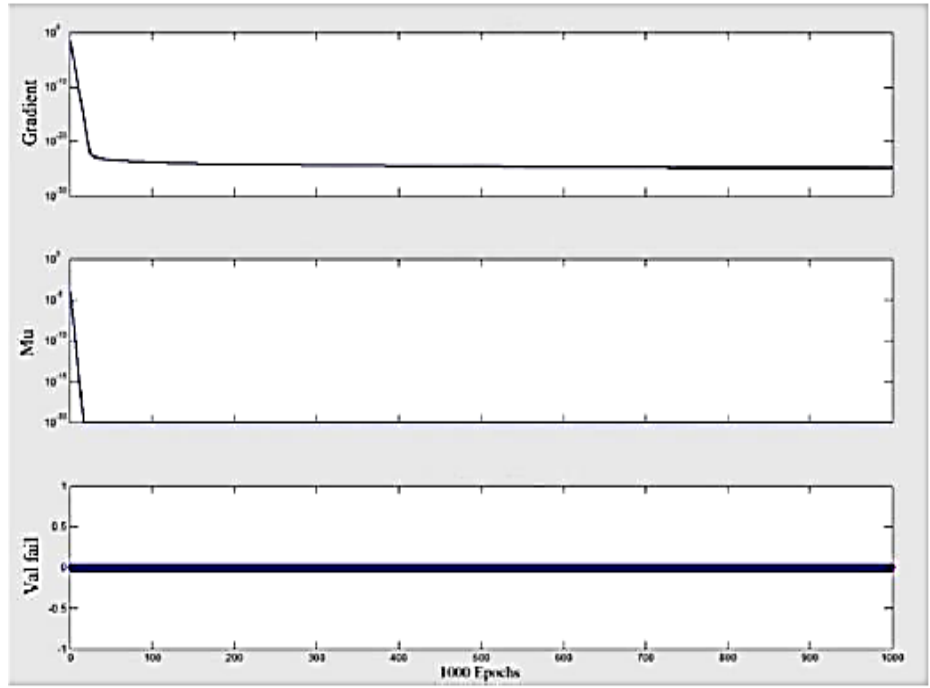

Figure 12: Training State Results

## 4. CONCLUSION

This manuscript introduces an Artificial Neural Network (ANN) derived approach for the diagnosis and classification of Anemia. Empirical findings illustrate the efficacy of the proposed methodology in adeptly identifying and discriminating among various types of Anemia. Conversely, the comparative assessment suggests superior performance of the Elman model, particularly noteworthy given its proficiency even with limited training data. Future endeavours may explore the amalgamation of image processing methodologies and ANNs, incorporating smear blood images to enhance the diagnostic and classificatory capabilities in the context of Anemia.

## ACKNOWLEDGEMENT

The authors express their genuine gratitude to the hematology unit at Omdurman Hospital, Sudan, for generously supplying a wide range of Anemia samples. They also extend heartfelt appreciation to the committed hospital staff and technicians for their invaluable aid and backing.

**AUTHOR**

In 2009, **Dr. Mohammed Elmaleeh** successfully completed his PhD studies and subsequently held positions as an assistant and associate professor in Sudan for several years. His academic passage led him to join the Computer Engineering Department at FCIT, Tabuk University in Saudi Arabia in September 2014. Dr. Elmaleeh boasts an extensive track record of publications in the realms of electrical and electronic engineering. He has provided mentorship to numerous postgraduate and undergraduate students in their research pursuits. Currently, he supervises the research projects of two PhD students specializing in control engineering. Dr. Elmaleeh has been recognized as a reviewer for various IEEE conferences and international journals. Furthermore, he has been selected as the chair for the control engineering track in multiple IEEE conferences. His research interests span a wide range, including embedded systems, control engineering,